\documentclass{article}

\usepackage{PRIMEarxiv}
\usepackage[utf8]{inputenc} 
\usepackage[T1]{fontenc}    
\usepackage{hyperref}       
\usepackage{url}            
\usepackage{booktabs}       
\usepackage{amsfonts}       
\usepackage{nicefrac}       
\usepackage{microtype}      
\usepackage{lipsum}
\usepackage{fancyhdr}       
\usepackage{graphicx}       
\graphicspath{{media/}}     
\usepackage{amsmath}
\usepackage{tikz}
\usepackage{bbm}
\usepackage{multirow}
\usepackage{xcolor}
\usepackage{colortbl}

\newcommand{\citet}[1]{\cite{#1}}

\title{A Collection of Deep Learning-based Feature-Free Approaches for Characterizing Single-Objective Continuous Fitness Landscapes
\thanks{Accepted at \textit{Genetic and Evolutionary Computation Conference (GECCO '22), July 9--13, 2022, Boston, MA, USA}\\
ACM ISBN 978-1-4503-9237-2/22/07...\$15.00 \\
\url{https://doi.org/10.1145/3512290.3528834}}
}

\author{
  Moritz Vinzent Seiler \\
  Data Science: Statistics and Optimization \\
  University of M\"unster, Germany \\
  \texttt{moritz.seiler@uni-muenster.de} \\
   \And
  Raphael Patrick Prager \\
  Data Science: Statistics and Optimization \\
  University of M\"unster, Germany \\
  \texttt{raphael.prager@uni-muenster.de} \\
  \And
  Pascal Kerschke \\
  Chair of Big Data Analytics in Transportation \\
  TU Dresden, Germany \\
  \texttt{pascal.kerschke@tu-dresden.de} \\
  \And
  Heike Trautmann \\
  Data Science: Statistics and Optimization \\
  University of M\"unster, Germany \\
  University of Twente, The Netherlands \\
  \texttt{heike.trautmann@uni-muenster.de} \\
}
\newcommand{\figcolumnwidth}{227pt}

\begin{document}
\maketitle

\begin{abstract}
Exploratory Landscape Analysis is a powerful technique for numerically characterizing landscapes of single-objective continuous optimization problems. Landscape insights are crucial both for problem understanding as well as for assessing benchmark set diversity and composition. Despite the irrefutable usefulness of these features, they suffer from their own ailments and downsides. Hence,  in this work we provide a collection of different approaches to characterize optimization landscapes. Similar to conventional landscape features, we require a small initial sample. However, instead of computing features based on that sample, we develop alternative representations of the original sample. These range from point clouds to 2D images and, therefore, are entirely feature-free.
We demonstrate and validate our devised methods on the BBOB testbed and predict, with the help of Deep Learning, the high-level, expert-based landscape properties such as the degree of multimodality and the existence of funnel structures. The quality of our approaches is on par with methods relying on the traditional landscape features. Thereby, we provide an exciting new perspective on every research area which utilizes problem information such as problem understanding and algorithm design as well as automated algorithm configuration and selection.

\end{abstract}



\maketitle

\section{Introduction}
The merits of a numerical characterization of single-objective continuous black-box optimization problems have been indubitably proven in various works. Especially, the use of \textit{Exploratory Landscape (ELA)} features in the areas of algorithm selection 
improved the state-of-the-art at that time significantly \cite{kerschke2019bbob, prager2020cc}.
However, as shown in the early days of ELA research, landscape features can also be used to classify a given problem instance w.r.t. its high-level properties \cite{Mersmann2011}. 
Hence, we deem it useful to revisit the work of \cite{Mersmann2011} from a different angle.


As Deep Learning (DL) has evolved into a highly competitive class of machine learning algorithms in the last decade, 
the potential of a feature-free landscape analysis arises. Thus, we will demonstrate in this work how to use the largely unprocessed fitness landscape information to accurately characterize fitness landscapes using DL, and skipping the intermediate step of calculating instance features as there are several drawbacks related to this process: the features (1) are designed manually in a tedious process, (2) require additional and (sometimes) computationally-expensive calculations, and (3) are tailored to a specific task or problem \cite{seiler2020deep}.
A first experimental study in the numerical domain has highlighted the potential of a so-called `fitness map' in combination with DL for automated algorithm selection \cite{PragerMP2021TowardsFeatureFree}. 
However, we enhance this approach and also provide further alternatives to the fitness map, which alleviates one of their largest drawbacks, i.e., we lift the restriction that the previously proposed fitness map is only applicable for $2d$ problems.

This paper is organized as follows. Section~\ref{sec:bg} provides a the used concepts and techniques throughout the paper. We detail various necessary concepts about the fitness landscape in Sections~\ref{sec:bb_landscape} and \ref{sec:highlevel}, followed by the basic notion of traditional landscape features, the fitness map, and the fitness cloud (Sections~\ref{sec:landscape-features}-\ref{sec:fcloud}). An experimental study\footnotemark is presented in Section~\ref{sec:setup_main} and its results are discussed in Section~\ref{sec:discussion}. Finally, Section~\ref{sec:conclusion} concludes our paper.
\section{Background}\label{sec:bg}

\subsection{Black-Box Problems and Fitness Landscapes}
\label{sec:bb_landscape}
\footnotetext{\url{https://github.com/Reiyan/highlevel\_property\_prediction.git}}

\begin{table}[!t]
	\centering
	\renewcommand{\arraystretch}{0.925}
	\caption{Characterization of the 24 BBOB functions based on their high-level properties multimodality, global structure, 
	and funnel structures. Note that this table combines information from tables provided in \cite{mersmann2010} and \cite{kerschke2015}.}
	\small
\begin{tabular}{@{}lrrr@{}}
	\toprule
	{\bf BBOB Function} & {\bf Multim.} & {\bf Global Str.} & {\bf Funnel}\\\midrule
    1: Sphere & none & none & yes \\
    2: Ellipsoidal separable & none & none & yes  \\
    3: Rastrigin separable & high & strong & yes \\
    4: B\"uche-Rastrigin & high & strong & yes \\
    5: Linear Slope & none & none & yes \\ \midrule
    6: Attractive Sector & none & none & yes \\
    7: Step Ellipsoidal & none & none & yes  \\
    8: Rosenbrock & low & none & yes\\
    9: Rosenbrock rotated & low & none & yes\\ \midrule
    10: Ellipsoidal high conditioned & none & none  & yes \\
    11: Discus & none & none & yes\\
    12: Bent Cigar & none & none & yes\\
    13: Sharp Ridge & none & none & yes\\
    14: Different Powers & none & none & yes\\ \midrule
    15: Rastrigin multimodal & high & strong & yes\\
    16: Weierstrass & high & med. & none\\
    17: Schaffer F7 & high & med. & yes\\
    18: Schaffer F7 moderately ill-cond. & high & med. & yes\\
    19: Griewank-Rosenbrock & high & strong & yes\\ \midrule
    20: Schwefel & med. & deceptive & yes\\
    21: Gallagher 101 Peaks & med. & none & none\\
    22: Gallagher 21 Peaks & low & none & none\\
    23: Katsuura & high & none & none\\
    24: Lunacek bi-Rastrigin & high & weak & yes\\\midrule
\end{tabular}
	\label{tab:bbob-functions}
\end{table}

In single-objective continuous optimization the aim is to find w.l.o.g. the global minimum of an objective or fitness function $f$, which maps decision variables $\vec{x} = (x_1,\ldots,x_d)$ to objective values $f(\vec{x})$ subject to constraints $g_i$ :
\begin{eqnarray*}
	\mbox{min.} \,\, f(\vec{x}) & &  s.t. \,\, g_i(\vec{x}) \le b_i \quad  (i=1,\ldots,k)\\
	f, g_i: & & \mathcal{X} \subseteq \mathbb{R}^d \rightarrow  \mathbb{R}
\end{eqnarray*}
Black-box optimization assumes that the exact equation of $f$ and thus the mechanism relating decision variables to objective values is unknown. Hence, algorithms, e.g., cannot exploit exact information on gradients and optimization becomes extremely challenging.

While a very preliminary landscape visualization 
w.r.t. gene combinations in evolution can already be found in \cite{wright1932}, we formally define a \emph{fitness landscape} $\mathbb{L}$ similar to \cite{muller2011global} as a triplet  $\mathbb{L} := (\mathcal{X}, d_E, f)$ 
with Euclidean distance $d_E$ between points, assuming a box-constrained decision space $\mathcal{X}$ with lower and upper bounds $[\mathbf{l},\mathbf{u}] \subseteq \mathcal{X}$. The landscape metaphor is herein used to characterize optima, plateaus and local structures in analogy to topographical structures in nature such as mountains, valleys, ridges, plateaus, lowlands, etc.

\subsection{High-Level Properties of Fitness Landscapes}
\label{sec:highlevel}
One of the key objectives when studying (single-objective continuous) optimization problems
is to get a better understanding of their structural characteristics. These insights, in turn, allow us (1) to find (dis)similarities between test problems, (2) to select and configure algorithms that have a higher chance of performing well on the given problem(s), and (3) to design new algorithms that are particularly good at dealing with challenging obstacles of the problem(s) under investigation.

As visual investigations of fitness landscapes are in most cases infeasible -- visualizations are mainly limited to problems with at most two input parameters -- we tend to help ourselves by describing (even high-dimensional) problems using rather tangible terms and concepts that are comparatively easy to understand, such as modality, plateaus, or funnel shapes (without seeing them). Most of the notions that we commonly use are not quantifiable by a single number, but instead describe the landscapes by means of high-level characteristics. 
For the 24 problems of the well-known BBOB test suite \cite{Hansen2009_Noiseless}, an overview of three high-level properties is given in Table~\ref{tab:bbob-functions}.
Other than these three,  additional five high-level properties are also commonly used in the \textit{Evolutionary Computation} (EC) community. 
These five properties are \textit{separability}, \textit{variable scaling}, \textit{search space homogeneity}, \textit{basin size homogeneity}, and  \textit{global to local optima contrast} \cite{Mersmann2011,mersmann2010,malan2009,kerschke2015}.
The three properties used throughout this paper are explicitly known to severely influence problem hardness and are summarized in the following: 
\begin{itemize}
\item The degree of \textit{multimodality} provides an aggregation of the problem's number of local optima.

\item The structural relationship of all local and/or global optima (i.e., ignoring all non-optimal points) is summarized within the problem's \textit{global structure}.






\item Considering both the number and the layout of the optima, the landscape of a problem can also be described in terms of whether its optima are aligned in a \textit{funnel}-shaped structure. A funnel exists, if the problem's optima pile up to an up-side-down version of a mountain.


\end{itemize}

%
%
\subsection{Landscape Features}
\label{sec:landscape-features}
\begin{figure*}[!h]
    \begin{minipage}{0\figcolumnwidth}
        \centering
        \begin{minipage}{0.48\linewidth}
            \centering
            \includegraphics[width=0.7\textwidth, trim={15mm 15mm 15mm 15mm}, clip]{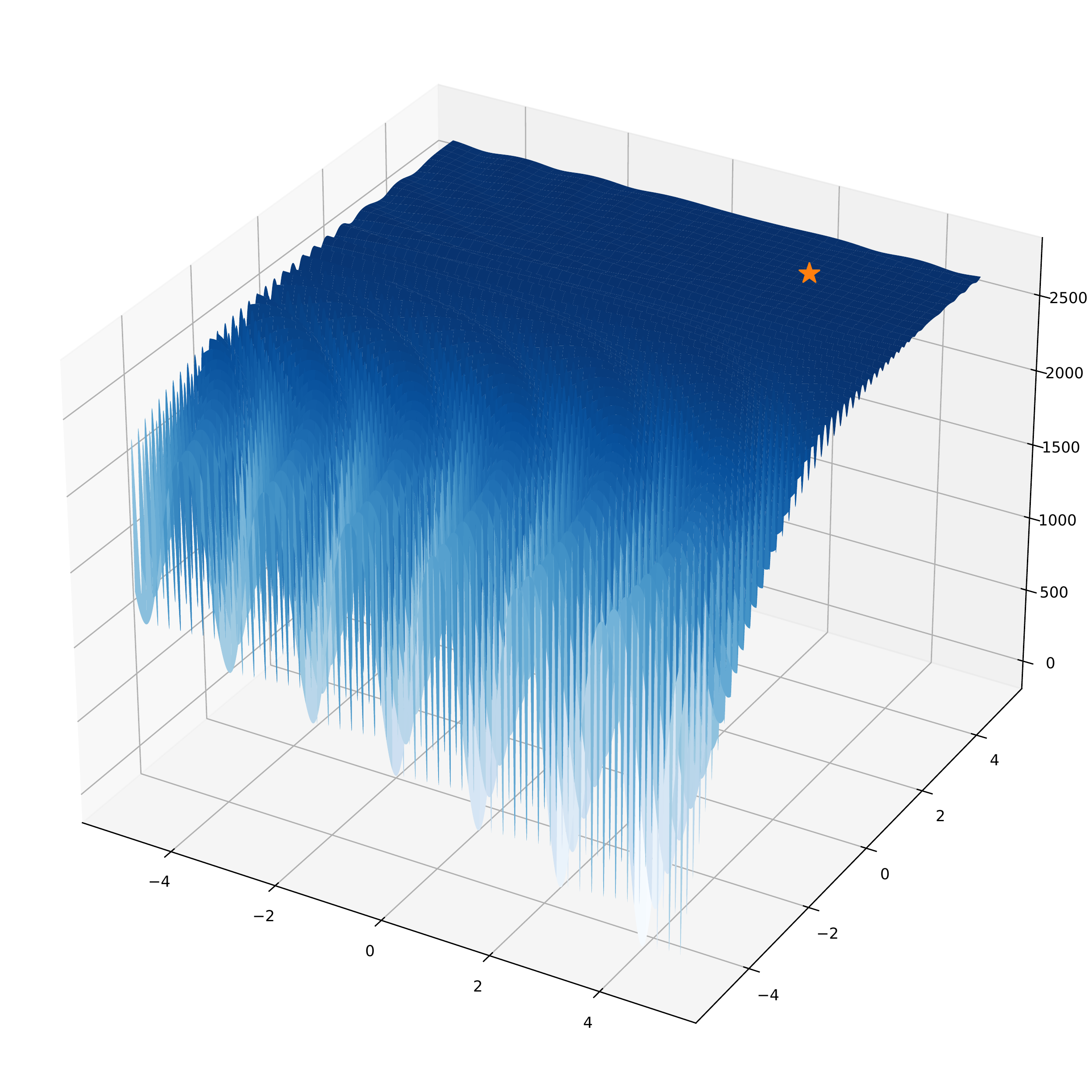}
            $z = f_{opt} - f(\vec x)$
        \end{minipage}
        \begin{minipage}{0.48\linewidth}
            \centering
            \includegraphics[width=0.7\textwidth, trim={15mm 15mm 15mm 15mm}, clip]{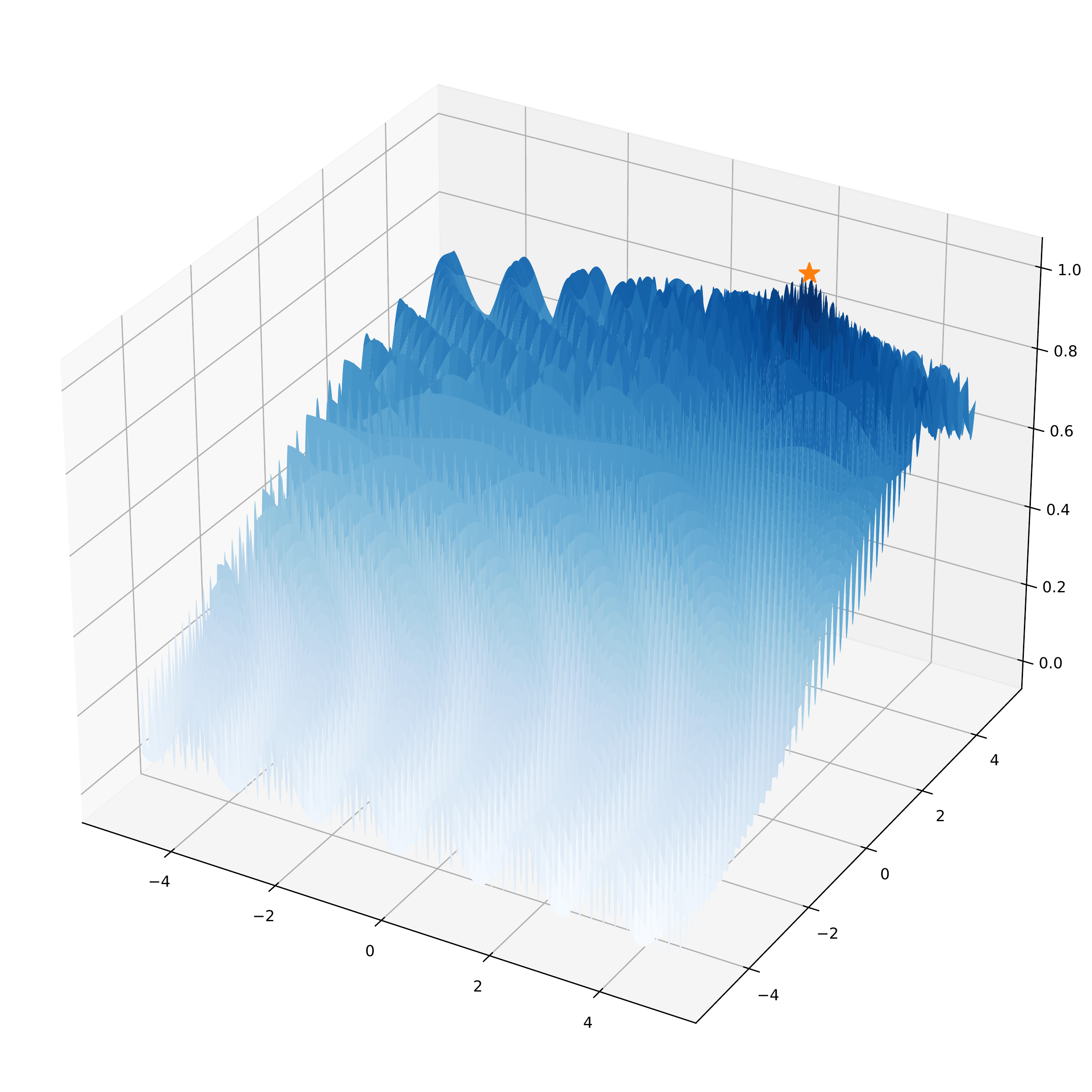}
            $z = 1 - \hat f(\vec x)$
        \end{minipage}
        
        \caption{Visualization of function 17 of the BBOB suite \cite{Hansen2009_Noiseless} in 2d as a max. problem. The position of the optimum $f_{opt}$ is highlighted in orange. The output is shown unnormalized (left), and normalized using Eq.~(\ref{eq:normalization_f}) (right).}
        \label{fig:f17}
    \end{minipage}
    \hspace{\columnsep}
    \begin{minipage}{\figcolumnwidth}
        \centering
        \includegraphics[width=0.65\linewidth, trim={40mm 18mm 40mm 18mm}, clip]{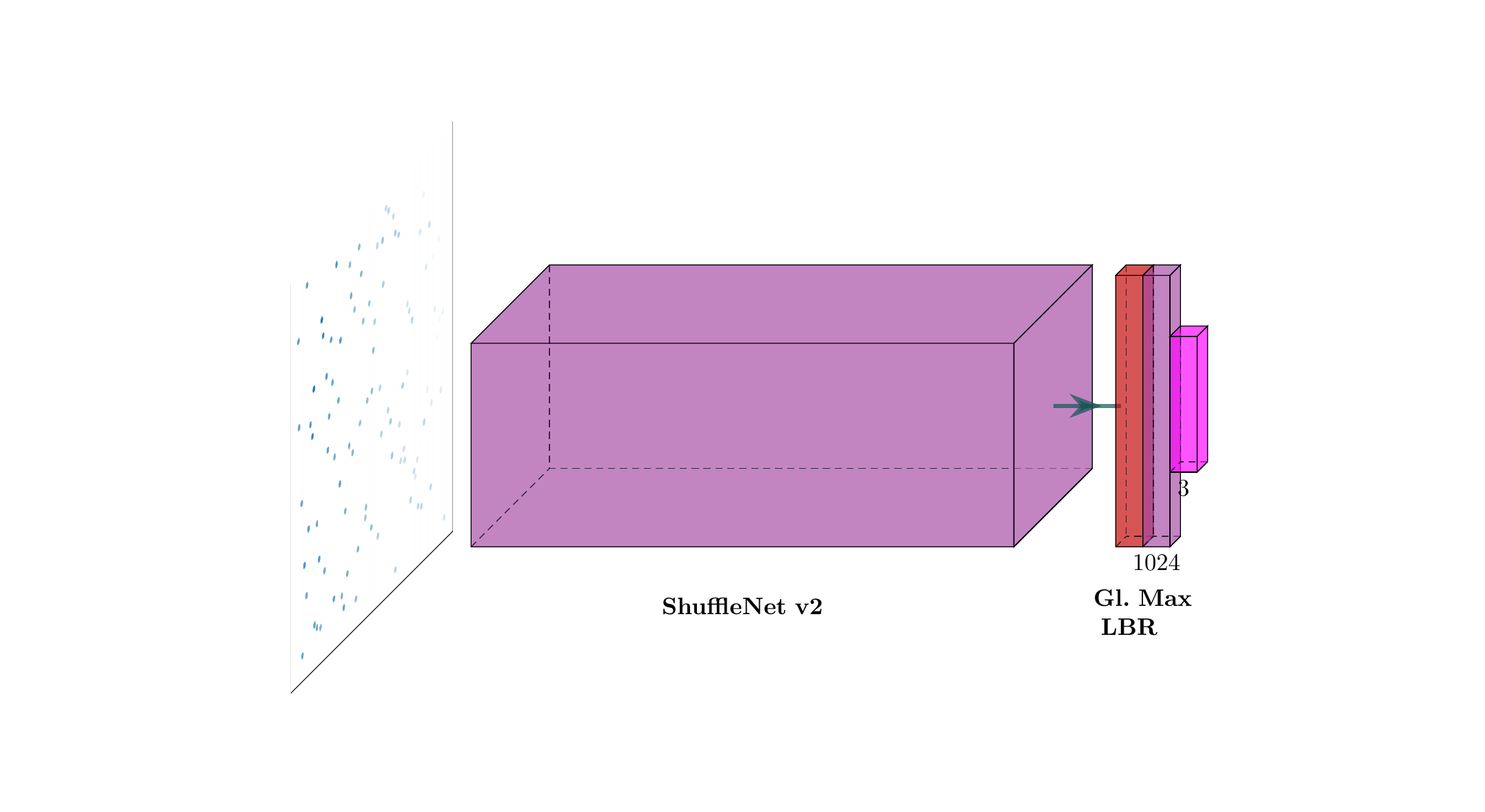}
        \caption{Our used Deep Learning architecture for the image-based approach. It consists of a ShuffleNet v2 \cite{wang2017cnn} with a Global Pooling layer + LBR (\textit{Linear, Batch-Normalization} \cite{ioffe2015batch}, and \textit{ReLU} \cite{nair2010rectified}).}
        \label{fig:cnn_fitness_map}
    \end{minipage}
\end{figure*}

As stated above, high-level properties can be very useful for describing a problem in a human-understandable way. However, characterizing new (i.e., previously unseen) black-box problems by means of those properties can be very challenging as they cannot be (directly) computed in an automated manner.
Instead, a property's attributes are usually assigned by human experts, which presents a severely limiting factor in automated processes such as automated algorithm selection. Another obstacle arises from the fuzzy transitions between adjacent attributes. For instance, there is no strict, separating boundary between a low and a medium, or a medium and a high degree of multimodality. 

With the help of low-level landscape features, the aforementioned obstacles can be overcome. These features are maximally informative, i.e., numerical values that can be calculated in an automated manner based on a small sample of (evaluated) points. Note that previous works have studied the impact of sample size \cite{kerschke2015} and sampling strategy \cite{bossek2020initial}; in fact, even the population of an evolutionary algorithm could be used as input for the feature calculation \cite{jankovic2021towards}.
By using a suitable combination of several of these features, it is in turn possible to draw conclusions about the high-level properties of the underlying problem. Moreover, as the features enable the (exploratory) analysis of the problem's fitness landscape, the corresponding line of research is dubbed \textit{Exploratory Landscape Analysis (ELA)} \cite{Mersmann2011} or sometimes also \textit{Fitness Landscape Analysis (FLA)} \cite{malan2013}.

Research in ELA has been on the rise for many years. As a result, hundreds of landscape features have been proposed over the last two to three decades. Hence, we herein refer the interested reader to the respective original references for a detailed description of these features (due to space limitations), and instead restrict ourselves to a brief listing of the considered feature sets:
\begin{itemize}
    \item \textit{Classical ELA}: While the complete set of ELA features covers six features sets, each containing multiple features, we focus on the three feature sets level set, meta model, and $y$-distribution (consisting of 22 features in total). The remaining feature are excluded because they require additional function evaluations to compute \cite{Mersmann2011}. 
    \item \textit{Nearest Better Clustering}: A small feature set (5 features), which summarizes the distances of a point to its nearest neighbor and its nearest better neighbor (for all points from the considered sample) \cite{kerschke2015}.
    \item \textit{Dispersion}: The 16 features of this set compare the spread of the distances of the better points to the spread of the distances of all points from the considered sample \cite{lunacek2006}.
    \item \textit{Information Content}: The 5 sequence-based features of this set rely on an enhanced version of the information content method from the combinatorial domain \cite{munoz2015_ic,vassilev2000}.
    \item \textit{Fitness Distance Correlation}: In total 6 features which capture the distance of the points in the decision space in relation to their respective objective values \cite{jones1995}.
    \item \textit{Miscellaneous}: Nine features which are based on principal component analysis and very basic meta information of the problem, such as its dimensionality \cite{kerschke2019flacco}.
\end{itemize}

\subsection{Fitness Map}
\label{sec:fmap}

Computer vision is one of the oldest and, at the same time, very well understood tasks of DL. \textit{Convolutional Neural Networks} (CNN) outperform humans in many visual tasks such as object detection and image segmentation (e.g. \cite{liu2020deep}). 
Therefore, it is consequential to explore the potential of using CNNs for landscape analysis based on images visualizing a small sample of points, 
also called \textit{Fitness Map}~\cite{PragerMP2021TowardsFeatureFree}. For every candidate solution, the coordinates $\vec{x}$ and the corresponding fitness-value $f(\vec{x})$ are known. Therefore, $X \in \mathbb{R}^{n\times d}$ contains the coordinates of a set of $n$ candidate solutions with dimensionality $d$ and $f(X)$ contains the fitness values of every candidate solution whereby $f(X) \in \mathbb{R}^n$.
We normalize both $X$ and $f(X)$ by
\begin{align}
        \hat{X} &= \frac{X - l}{u - l + \epsilon} \label{eq:normalization_x} \\[0.25em]
        \hat{f}(X) &= \frac{\ln\big[1 + f(X) - \min(f(X))\big]}{\ln\big[1 + \max(f(X)) - \min(f(X))\big] + \epsilon}
        \label{eq:normalization_f}
\end{align}
where $\epsilon=10^{-8}$ is a very small number to countervail division by zero and $[\mathbf{l},\mathbf{u}] \subseteq \mathcal{X}$ because we assume box-constraints for $X$ as explained in Section \ref{sec:bb_landscape} (see Fig.~\ref{fig:f17} for an exemplary normalization). Afterwards, $\hat X \in [l,u]$ and $\hat f(X) \in [0,1]$. Normalization is important for the image-based approaches because the image's resolution as well as the color range for each pixel are finite.
To generate the image or \textit{fitness map} for an instance, the fitness values $\hat{f}(X)$ are mapped into a Cartesian plane at the locations given by $\hat X$. Afterwards, the plane is converted into a gray scaled image and the gray values represent the fitness value. Unknown areas without any sampled solutions are left white as originally proposed in \cite{PragerMP2021TowardsFeatureFree}.

\begin{figure*}
    \centering
    \input{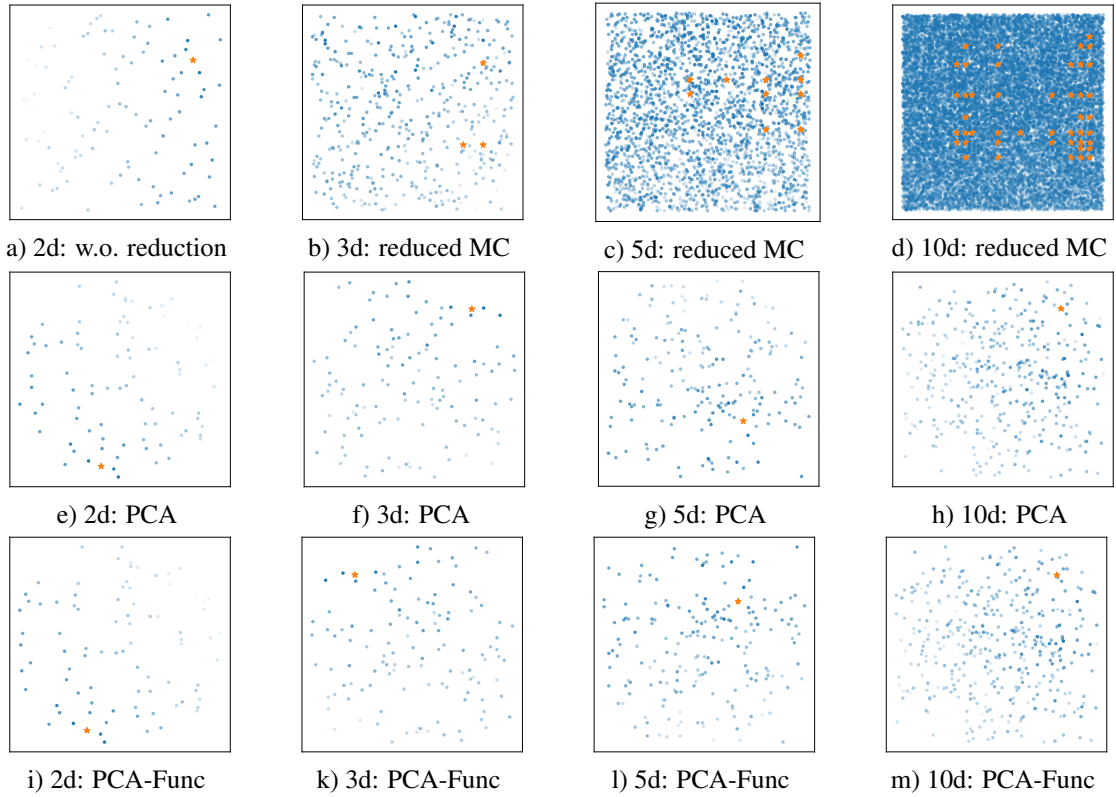}
    \caption{Visualization of the fitness map approaches with different reduction techniques of F17. The full decision space can be found in Figure \ref{fig:f17}. The optimal value $f_{opt}$ is shown in orange for better reference but left out in the original images. $y$ is normalized by Equation \ref{eq:normalization_x} and \ref{eq:normalization_f}. Note: (1) points have been increased in size for better visualization, in the original fitness map every point is only 1px in size; (2) for the rMC approach, $f_{opt}$ is shown several times from different perspectives.}
    \label{fig:reduction_methods_f17}
\end{figure*}

Every convolutional layer has a weight matrix, called \textit{convolution-kernel} or just \textit{kernel}. The input image is convoluted by the trainable kernel into an output image. However, one major constraint of CNNs is their limitation to two or three dimensional data as the number of weights in each convolution-kernel 
grows exponentially with the number of dimensions. Therefore, we extended the approach of \cite{PragerMP2021TowardsFeatureFree} by introducing four different dimensionality reduction techniques to reduce the fitness map's dimensionality to two and, thus, enable 2d-CNNs to compute high-dimensional landscapes, efficiently.

\subsubsection{Principal Component Analysis (PCA) \cite{pearson1901lines}} This is the first technique that we used for reducing the dimensionality of $X$. To do so, we applied PCA to $X$ without centering the data a priori to keep the location of the points in the same trajectory. In detail, we performed the following steps 
\begin{align}
      \Tilde{X} &= \lambda_{1,2}^{T} \cdot X; \; \:\lambda_i = var(S); \; \:S = cov(X^T),
\end{align}
where $cov$ is the covariance matrix of $X^T$ and $\lambda_i$ are all eigenvectors of $S$. $\Tilde{X} \in \mathbb{R}^{n\times 2}$ is then normalized as decribed in Eq. (\ref{eq:normalization_x}).
Points that exceed the upper and lower bounds are dropped (which is rarely the case) to keep the resolution of the fitness map consistent; 
varying resolutions of the fitness maps may impact a meaningful representation of distances between neighboring points. As explained previously, 
the fitness values $\hat{f}(X)$ 
are then mapped into a Cartesian plane at the locations $\Tilde{X}$.
        
\subsubsection{Principal Component Analysis incl. Fitness Value (PCA-Func)} 
In contrast to the former approach, this one also considers $\hat{f}(X)$ to find more meaningful \textit{principal component}s (PC) w.r.t. $f(X)$.
Similarly, we performed the following steps 
\begin{align}
    \Tilde{X} = \lambda_{1,2}^{T} \cdot \acute X; \;\: \lambda_i = var(S); \;\: S = cov(\acute X^T); \;\: \acute X = X \: || \: \hat{f}(X)
\end{align}
where $||$ is the operator for concatenation and $\acute X \in \mathbb{R}^{n\times (d+1)}$. All other steps are identical to the default PCA procedure. Afterwards, the fitness values $\hat f(X)$ 
are mapped in the same way as previously described into a Cartesian plane.

\subsubsection{Multiple Channel (MC)} This approach follows the idea of Liu et al.~\cite{liu2019relation}. For each possible pairwise combination of decision variables in $X$ an individual fitness map is generated. Note that the total number of all fitness maps corresponds to $c=\binom{d}{2}$, e.g., for $d=5$ there exist 10 fitness maps while for $d=10$ a total of 45 different fitness maps are created. These fitness maps are then stacked into an image with $c$ channels. Usually, CNNs are designed to work with RGB images and, thus, expect a three-channel input. Yet, the number of expected input channels can be easily adjusted to any finite number. However, one major drawback of this approach is that $d$ is limited by the number of available input channels $c$. For instance, if $c=45$, the dimensionality $d$ of $X$ is limited to 10 (at most), because the number of possible pairwise combinations would otherwise exceed the number of input channels $c$. 
However, if $X$ has less dimensions, the missing channels can be filled with white or empty fitness maps.

\subsubsection{reduced Multiple Channel (rMC)} Our last proposed image-based approach aims to overcome the limiting factor of our MC approach. The $c$ fitness maps are generated in the same manner as in the MC approach. Afterwards, they are reduced to a single fitness map by simple \textit{mean}-aggregation excluding empty cells. 
We expect that this approach works great for smaller $d$, but may suffer for larger $d$, because the model cannot distinguish between the numerous 2d-projections.\\[0.1em]

Fig.~\ref{fig:reduction_methods_f17} shows examples for three of the four dimensionality reduction methods. The MC approach is missing as it is impossible to visualize images with more than three color-channels. The image-based approaches have their benefits and drawbacks. On the one hand, CNNs are -- in comparison to other, newer fields in DL -- well understood and, therefore, training is (in-comparison) easy.
On the other hand, images may not be an ideal representation of a set of candidate solutions as the exact location of each point is lost. This is due to the fact that an image consists of a finite number of pixels and each pixel represents a specific cell within the Cartesian plane. Therefore, the resolution of the coordinate system depends on (1) the number of available pixels, and (2) the upper and lower bounds of the fitness map. In addition, all image-based approaches, except for MC, loose information for $d>2$ for every point as dimensionality reduction methods cause an information loss. Yet the MC approach has its limitations as well. As the number of channels grows with $c=\binom{d}{2}$, the number of channels and weights within a CNN grow rapidly for high dimensions. 

\subsection{Fitness Clouds}
\label{sec:fcloud}
\begin{figure*}[h]
    \centering
    \begin{minipage}{0.33\textwidth}
        \centering
        \begin{tikzpicture}[scale=0.95]
    \filldraw[draw=black, fill=none, anchor=west] (2.5,4) -- (2.5,8.5) -- (7,8.5) -- (7,4) -- (2.5,4);

    \node[circle,fill=black,inner sep=0pt,minimum size=5pt,label=below:{$\vec x_i$}] (a) at (5,5) {};
    
    \node[circle,fill=blue,inner sep=0pt,minimum size=5pt,label=below:{$\vec x_{j=1}$}] (a) at (6,5) {};
    \node[circle,fill=blue,inner sep=0pt,minimum size=5pt,label=below:{$\vec x_{j=2}$}] (a) at (3.4,4.8) {};
    \node[circle,fill=blue,inner sep=0pt,minimum size=5pt,label=right:{$\vec x_{j=3}$}] (a) at (3,7.1) {};
    \node[circle,fill=red,inner sep=0pt,minimum size=5pt,label=left:{$\vec x_{j=4}$}] (a) at (6.5,8) {};
    
    \draw [-stealth, dashed, draw=blue](5.15,5) -- (5.85,5);
    \draw [-stealth, dashed, draw=blue](4.85,4.95) -- (3.55,4.65);
    \draw [-stealth, dashed, draw=blue](4.9,5.1) -- (3.1,6.9);
    \draw [-stealth, dotted, draw=red](5.05,5.15) -- (6.45,7.9);
    
\end{tikzpicture}\\
        (a)
    \end{minipage}
    \begin{minipage}{0.66\textwidth}
        \centering
        \includegraphics[width=0.9\linewidth, trim={0mm 0mm 0mm 0mm}, clip]{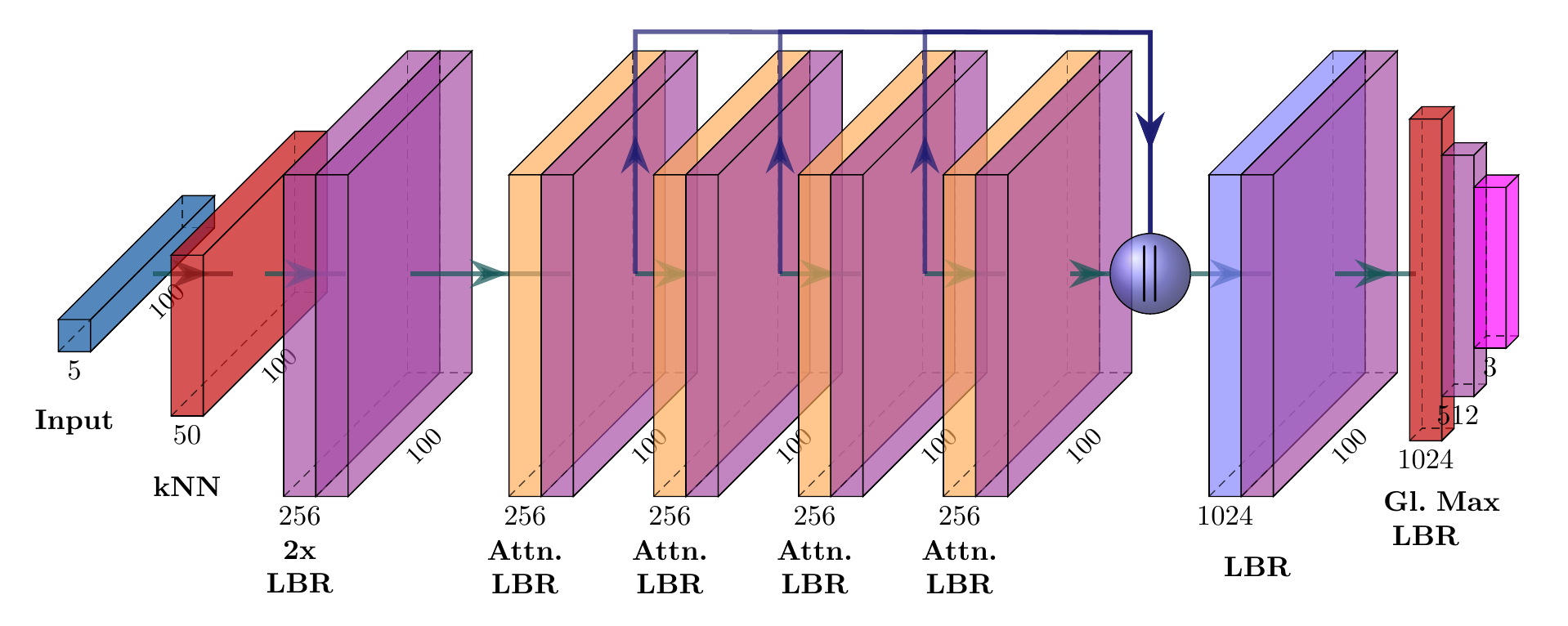}\\
        (b)
    \end{minipage}
    \caption{Visualization of the transformer model including the embedding. (a) for every $x_i \in X$ the $k-1$NN are calculated. If the distance between $\vec x_i$ and $\vec x_j$ exceeds a certain limit $\Delta_{max}$ (shown in red), the point $\vec x_j$ is replaced by $\vec x_{j-1}$ to preserve local neighborhood \cite{shaw2018self}. The output for this case would be $\vec{x}_j = \vec x_i \:||\: \vec x_{j=1} \:||\: \vec x_{j=2} \:||\: \vec x_{j=3} \:||\: \vec x_{j=3}$. (b) The adapted architecture as proposed by Guo et al.~\cite{guo2021pct}, but with a different input-embedding using more features in the attention layers (Attn.), and with only two final linear layers. LBR stands for \textit{Linear, Batch-Normalization} \cite{ioffe2015batch}, and \textit{ReLU} \cite{nair2010rectified}, GL Max for \textit{Global Max-Pooling}.}
    \label{fig:transfomer_fitness_cloud}
\end{figure*}
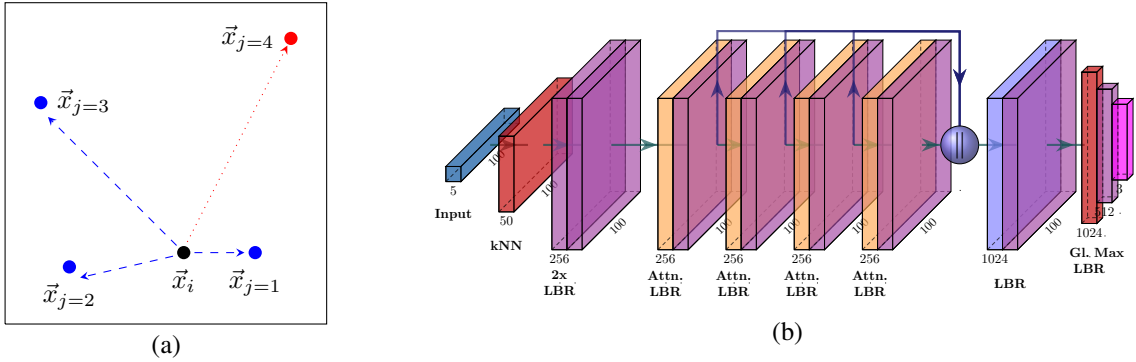

To countervail these shortcoming, we explored a second and novel field in DL, called \textit{3d-point cloud} analysis. DL for 3d-point clouds is most commonly used in the context of \textit{Light Detection and Ranging} (LiDAR) data \cite{li2020deep}. 
One of its main challenges is the \textit{point order isomorphism}:
the order of 3d-points within a set of points has no semantic meaning. Therefore, all operations within a DL model must respect the isomorphic nature of the input data: changing the order of the points must not influence the computed output. 

However, the fundamental operation in every DNN is $ y(\vec{x}) = \vec{w}^T \cdot \vec{x} + b $. If the order of $x_i \in \vec{x}$ changes, $y$ will change accordingly because the order of $w_i \in \vec{w}$ does not adapt automatically. Therefore,  operations in DNN that respect \textit{point order isomorphism} let the order of $w_i$ depend on the order of $x_i$ in a way that $y$ remains the same if the order of $x_i$ changes. Yet, these operations are not trivial and increase complexity of DNNs, substantially.

Several 3d-point cloud approaches have been proposed recently. Yet, most of these approaches operate on graph-like structures such as $kp$-graphs \cite{thomas2019kpconv}, \textit{k-nearest-neighbor} (kNN) graphs \cite{wang2019dynamic}, or some form of hierarchical pooling \cite{qi2017pointnet,wang2017cnn}. We assume that these methods may focus too much on a local neighborhood and, thus, may suffer from finding  meaningful global representations. Therefore, we adapted the idea of \textit{Point Cloud Transformers} (PCT) \cite{guo2021pct}.


As we want to predict \textit{high-level properties} such as \textit{multimodality} or \textit{global structure}, a method focusing solely on a local neighborhood may fail to identify these properties due to the fact that those rather depend on a global context than on local neighborhood. Another advantage of transformer for 3d-point cloud analysis is the fact that the attention mechanism within each layer operates in an isomorphic manner (see~\cite{guo2021pct} for more details). 
However, this may have its downsides as the attention mechanism might ignore positional relation between input data. Therefore, \cite{shaw2018self} introduced \textit{Relational Positional Representation} which we adapted for our own proposed \textit{embedding} strategy. 

PCT were originally proposed with a different form of embedding than we used in the context of this paper. \cite{guo2021pct} adapted the idea of \textit{Edge-Convolutions} \cite{wang2019dynamic} for their embedding strategy. These forms of convolutions operate on the edges within a $k$NN-graph. We did not find this process intuitive for our task of landscape analysis as edges within an artificially generated graph may not contain any useful information about the underlying landscape. In addition, there are significantly more edges in a $k$NN-graph (for $k>1$) as there are nodes, which increases computational complexity noticeably.

So, we propose our own approach of input embedding which does not operate on edges but on the nodes of the $k$NN-graph. 
Or, in other words, every $x_i \in X$ is embedded into its local neighborhood. So, we define $X_j$ as the set of all $x_i$ together with its $k-1$ nearest neighbors 
\begin{align}
    \hat X_j:& \: \mathbb{R}^{n\times d} \: \xrightarrow[]{kNN} \: \mathbb{R}^{n\times d \cdot k}; \quad 
    \hat f_j(x): \: \mathbb{R}^{n} \: \xrightarrow[]{kNN} \: \mathbb{R}^{n\times k} \\
    X_{Emb} &= \hat X_j \: || \: \hat f_j(X) \: || \: \mathbb{1}; \quad\quad\;\: \mathbb{1} \in \mathbb{R}^{n\times d}
\end{align}
where $X_{Emb} \in \mathbb{R}^{n\times k \cdot (d+1) + d}$ is the resulting embedded input and the $k$NN-graph applied to $\hat f(X)$ is identical to $X$.
To enable the embedding process to work with $X$s of different dimensionality, all $X$s with lower dimensionality receive additional coordinates with zeros to fill the missing dimensions. However, as zeros are valid coordinates and, thus, the model could not distinguish between the original and the appended dimensions, an additional indicator $\mathbb{1}$ is concatenated to $\hat X_j$ (together with $\hat f_j(X)$) to indicate which of the coordinates are valid or artificially appended.


We investigated the effects of $k\in\{1,3,5,10\}$ to find a suitable value for $k$, as well as the effects of different $\mathbb{L}_p$-norms as 
distance function for finding the $k$NN-graph. Often, the $\mathbb{L}_2$-norm (i.e., \textit{Euclidean} norm) is used to find the $k$NN-graph in 2d or 3d data. However, as the $\mathbb{L}_2$-norm is not well-suited for high-dimensional data, we also considered different $p \in \{1, 2, \infty \}$ as distance function. Next, as proposed by \cite{shaw2018self} we include a third parameter, $\Delta_{max}$ that limits the maximum distance in the $k$NN-graph search to preserve local neighborhood. If the distance between $x_i$ and $x_j$ exceeds $\Delta_{max}$, $x_j$ is replaced by the values of $x_{j-1}$. See Fig.~\ref{fig:transfomer_fitness_cloud} for an example.

In a small ablative study, we trained  various models with different $k, p, \text{and } \Delta_{max}$ on the training set and evaluated these models on the validation set. Based on this results, we found that $k \in \{3,5\}$, $p=\infty$ and $\Delta_{max}=1.5$ are particularly well-suited for our use case (see Section \ref{sec:setup_fcloud}). However, we found that $k$, $p$ and $\Delta_{max}$ in comparison to other hyperparameters 
(such as the number of layers, number of \textit{hidden} features, etc.) have a low impact on the overall performance of the transformer models. We believe, that $p=\infty$ works well on our task because it is unaffected by the number of dimensions. However, in future work we want to investigate the effect of $p$ more closely and, even, consider $p<1$ as proposed by \citet{aggarwal2001surprising}.


\section{Experiments}
\label{sec:setup_main}
Within our experiments, we trained several machine learning models to classify a set of benchmark problems w.r.t.~the three high-level properties \textit{multimodality}, \textit{global structure}, and \textit{funnel structure} (cf. Tab.~\ref{tab:bbob-functions}). We focused on these three properties because their degree of existence often determines an optimization problem's difficulty. 
This aforementioned set of benchmark problem is the well-known Black-Box Optimization Benchmark (BBOB) \cite{Hansen2009_Noiseless}. It is part of the \textit{COmparing Continuous Optimisers (COCO)} platform which nowadays covers a wider range of problem types, including (but not limited to) single-, multi-objective and noisy problems \cite{hansen2021}. The considered single-objective BBOB test suite is constituted of $24$ noiseless functions $F = \{1, 2, ..., 24\}$. These functions are organized into five groups, where each function group focuses on certain problem characteristics. For instance, the fourth and fifth group (i.e., F15 to F19 and F20 to F24, respectively) both consist of multimodal functions, but can be distinguished by their degree of global structure. 
Furthermore, each function comprises an infinite number of instances $I$, which maintain the function's general high-level properties, but offer a wider variety of optimization problems by means of shifts, rotations and scaling. Lastly, all the functions are of arbitrary dimensionality~$d$.

In our study, we considered all 24 BBOB functions, with the instances $I = \{1, \ldots, 150\}$, and dimensionality $D = \{2, 3, 5, 10\}$. This amounts to $24 \cdot 4 \cdot 150 = 14\,400$ distinct problems in total. Since the considered features and feature-free approaches are based on a randomly generated sample, we repeat the sample generation ten times per problem. This should help to a certain extent to account for the fitness map's, fitness cloud's, and ELA feature's variability (caused by the sample generation). 
In the following, the term \textit{problem instance} $\mathbf{p}$ is used as short form of the tuple $\mathbf{p}: = (f, d, i, r)$, where $f \in F$ and $i \in I$ are the function and instance ID of the BBOB test suite, $d \in D$ is the problem dimensionality, and $r$ is the current repetition.

We divide the set of $144\,000$ problem instances (including the 10 repetitions per $\mathbf{p}$) into three subsets: training, validation and test set. To give a realistic impression of our models' performance, we restrict the training, hyperparameter tuning and feature selection (in case of models based on ELA features) to the training and validation set and only afterwards assess the quality of a model on the test set. 
Thereby, we ensure that hyperparameter tuning and feature selection does not bias the model towards the test set.
The training set consists of all BBOB  problems of all dimensions but with instance ids only in the range of $\{1, \ldots, 100\}$. Following this scheme, the validation set consists of problem instances with instance ids in the range $\{101, \ldots, 125\}$, and the test set with instances in the interval $\{126, \ldots, 150\}$. We leave instances instead of entire functions out  because some high-level properties are very underrepresented and by removing an entire BBOB function from the training data set, we essentially set our models up for failure. The extreme case for this example is BBOB Function 20 which is the sole member of the class \textit{deceptive} of the high-level property global structure.

\subsection{Conventional ELA Feature Approach}
\label{sec:setup_ela}
Formally, we can define the training data set as $\mathbf{Z}_{Ela} = (X_{Ela}, Y)$, where $X_{Ela} \in \mathbb{R}^{n\times m}$  is a collection of \textit{input data}, $Y \in \mathbb{R}^{n\times 3}$ is the set of class labels for the three considered high-level properties $\vec{y}$, $n$ is the number of samples, and $m$ is the number of features which are used to classify an observation. In the scope of this work, $X_{Ela}$ is the set of considered landscape features and $\vec{y}$ the high-level property in question. In our experiments, we considered $m = 62$ landscape features for a total of $n = 14\,400$ observations. We create our training data $X_{Ela}$ by first generating a sample (using Latin Hypercube Sampling) on a problem instance $\mathbf{p}$, and subsequently calculating the features using the latest version of Python package \texttt{pflacco}\footnote{https://github.com/Reiyan/pflacco\_experiment}. 14 features frequently suffered from missing values. These predominately belong to the feature classes \textit{ela\_level} and \textit{dispersion}.
As there is no simple mechanism to deal with these values without introducing significant drawbacks, we eliminated these features from further consideration.

The three-dimensional class label $Y$ leads to a multilabel classification task, i.e., we can associate three class labels to each observation. A concept known to be able to deal with such tasks is called \textit{Binary Relevance (BR)} \cite{zhang2018binary}. Within BR, each label receives its own independent base classifier. For this, we considered \textit{Random Forest (RF)}, \textit{Gradient Boosting Trees (GBT)}, and \textit{Support Vector Machines (SVM)} which are all realized with the Python Package \texttt{sklearn} \cite{scikit-learn}.

To improve each base classifier, we performed hyperparameter tuning and feature selection on the validation set. For this very basic hyperparameter tuning, we employed a randomized search over the parameter space with five samples for each model. While this number can be perceived as low, the amount of data, the considered models as well as feature selection required an already extraordinary amount of computational resources which forced us to limit the number of sampled hyperparameter combinations. Dependent on the model, 
the considered parameters were (GBT) \texttt{n\_estimators}, \texttt{learning\_rate}, \texttt{max\_features}, (RF) \texttt{n\_estimators}, \texttt{criterion}, \texttt{max\_features}, (SVM) \texttt{C}, and \texttt{gamma}.

After obtaining the $X_{Ela}$ data set, we performed feature selection to eliminate features that either do not have any inherent predictive power (for the considered high-level properties), or do not provide additional information, which has not been previously captured by any of the features already included in the model. The purpose is to reduce the amount of non-discriminating and highly correlating features as a model's performance is susceptible in a negative way to such. For this purpose, we used the Python package \texttt{mlxtend} to perform a \textit{greedy forward-backward selection} \cite{raschkas_2018_mlxtend}.

Our final BR model consists of two GBT models for the high-level properties \textit{multimodality} and \textit{global structure}, whereas \textit{funnel structure} is captured by an RF model. 
For both GBT models, the best performance is achieved with $\texttt{n\_estimators} = 222$, $\texttt{learning\_rate} = 0.1389$, and $\texttt{max\_features} = \sqrt{m}$.
The hyperparameters of the RF model are $\texttt{n\_estimators} = 87$, $\texttt{criterion} = \text{`entropy'}$, and $\texttt{max\_features} = \sqrt{m}$. On the other hand, the process of feature selection eliminated six features across all feature sets.

\subsection{Fitness Map Approach}
\label{sec:setup_map}
For the image-based approaches, we used a \textit{ShuffleNet V2}~\cite{ma2018shufflenet} because of its efficient design with a $1.5\times$ width-multiplicator, identical to the model used by \cite{PragerMP2021TowardsFeatureFree} 
(see Fig.~\ref{fig:cnn_fitness_map}). In comparison to other architectures, the ShuffleNet V2 was designed to optimize FLOPS and memory usage. Thereby, it reduces the training time significantly, and still offers enough parameters to learn complex visual tasks. In contrast to the conventional approach (see Section \ref{sec:setup_ela}), a single model is used to predict all three considered high-level properties at once which we call \textit{multi property-prediction}. We performed an ablative study to measure the effect of the multi property-prediction (in-comparison to a single property-prediction) on the overall performance and found that this has no measurable impact. Hence, we assume that the \textit{ShuffelNet V2} has enough parameters to learn a well-suited internal representation to cover all three high-level properties, simultaneously. 

Similar to the previous section, we used $\mathbf{Z}_{Map} = (X_{Map}, Y)$ as our training set. We considered all four proposed dimensionality reduction methods and trained one model each across all dimensions~$D$. For the methods PCA, PCA-Func and rMC only a single input channel with a fitness map resolution of $224\times 224$ pixel is used, or $X_{Map} \in \mathbb{R}^{n\times 224 \times 224 \times 1}$. Yet, as the number of input channels varies depending on the dimension $d$ produced by the MC approach, we changed the default  \textit{ShuffleNet V2} to 45 input channels due to the fact that the MC method produces a fitness map with $45$ channels for $d=10$. For lower dimensions, we filled the missing channels with additional channels that contain only zeros. In this case, the input data is of shape: $X_{Map} \in \mathbb{R}^{n\times 224 \times 224 \times 45}$.

Next, we used \textit{ADAptive Moment estimation} (Adam) with Weight Decay \cite{loshchilov2017decoupled} as optimizer, $\lambda=5\cdot10^{-5}$ as learning rate and a batch-size of $32$ for training and validation as these are commonly used hyperparameters. 
All models were trained for $100$ epochs on a single Nvidia Quadro RTX 6000. During training, we used the validation set to measure overfitting and saved the best model based on the validation set. Afterwards, we evaluated the trained models on the test set as defined above. 

\subsection{Fitness Cloud Approach}
\label{sec:setup_fcloud}
For the transformer-based approach on 3d-point clouds, we trained  four transformer models in total. The models are identical to one another except for the input embedding and the number of sampled points. We doubled the number of hidden features 
in each attention layer, and used only a single \textit{Linear}, \textit{Batch-Normalization}~\cite{ioffe2015batch}, and \textit{ReLU}~\cite{nair2010rectified} (LBR) after the \textit{Global Pooling} layer as compared to the original proposed method \cite{guo2021pct}. After some initial testing, we found that this setup worked better for our case. However, we used our proposed input embedding strategy with two different configurations: $k \in \{3,5\}$, $p = \infty$, and $\Delta_{max} = 1.5$. Due to some framework limitations, we were not able to train the model with a variable size of sampled candidate solutions as we have used for the other two approaches. Instead, we had to use a fixed number of sampled points. In order to find a comparable setup, we trained both setups with $100$ and $500$ sampled points for every training instance, providing us with an upper and lower boundary of a model that could have been trained with a variable number of points.

The training data is of two different shapes: $X_{Cloud} \in \mathbb{R}^{n\times 100\times 10}$ and $X_{Cloud} \in \mathbb{R}^{n\times 500\times 10}$, depending on the number of sampled points. As explained in Section \ref{sec:fcloud}, for lower dimensional problems, the missing coordinates are filled with zeros. An additional indicator is provided to indicate which of the dimensions are valid. Therefore, the training set is $\mathbf{Z}_{Cloud} = (X_{Cloud}, f(X), \mathbb{1}, Y)$. Other from that, we used an identical setup for training as described in Section \ref{sec:setup_map}.

\section{Discussion}
\label{sec:discussion}

\begin{figure}[!t]
    \centering
    \includegraphics[width=\linewidth]{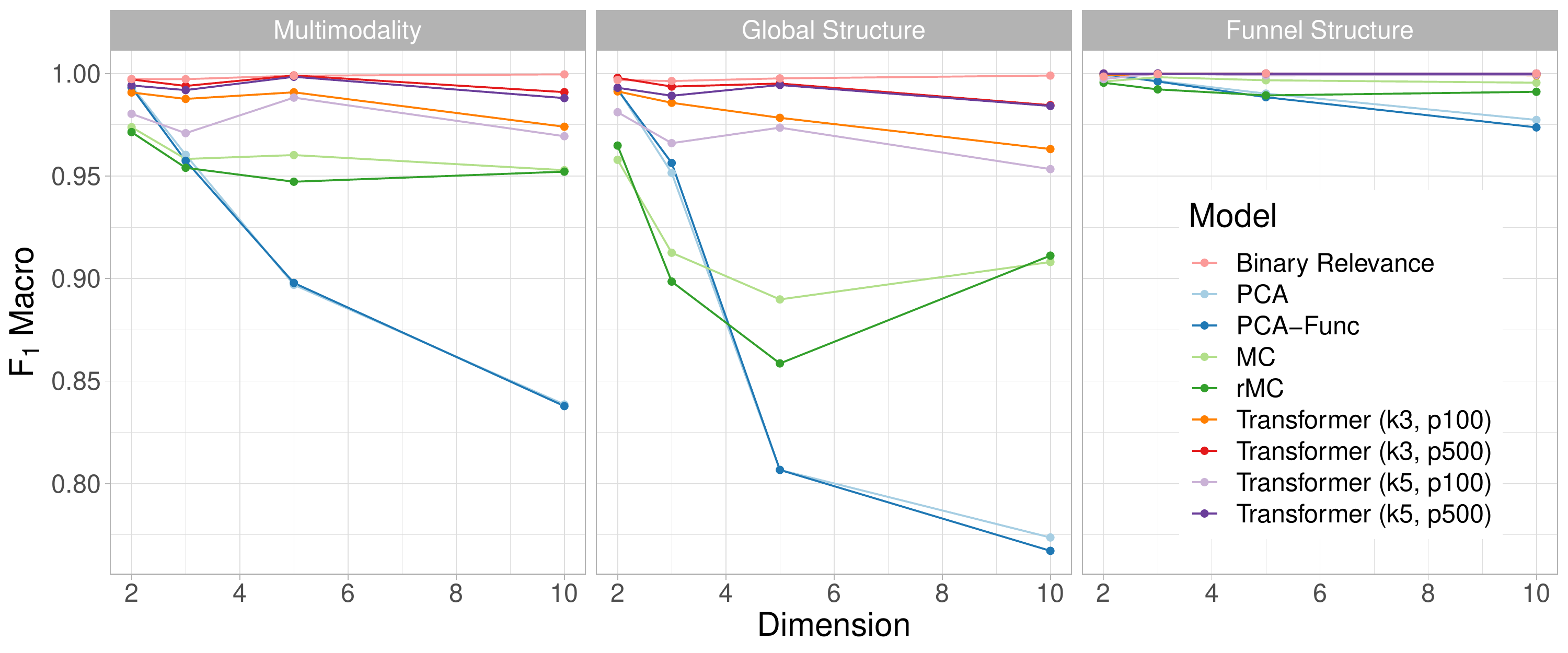}
    \caption{Visualization of the performance result divided by the properties and dimension. The used metric is $F_1$ Macro.}
    \label{fig:performance_lineplot}
\end{figure}

We considered the $F_1$-Score as our main performance measure as the classes within each high-level property are highly imbalanced. In contrast to \textit{funnel structures}, \textit{multimodality} and \textit{global structure} contain multiple classes which requires some form of aggregation. We decided on \textit{macro-averaging} within the $F_1$-score as it treats all classes equally by unweighted averaging of precision and recall for all classes 
and, thereby, counteracting the class-imbalance \cite{sokolova2009systematic}. The final results can be found in Tab.~\ref{tab:performance_results}. It becomes evident that the conventional approach marginally outperforms our proposed techniques in most settings. However, the image-based approaches suffer a noticeable drop in performance for $d\geq 3$. 
The drop in performance is the most severe for the two high-level properties \textit{multimodality} and \textit{global structure} (see Figure \ref{fig:performance_lineplot}). 

\begin{table}[!t]
	\centering
	\caption{Performance results of each used approach divided by high-level property and dimension. The listed value shows the $F_1$ Macro metric. Cells which are highlighted in grey, represent the best performing approach for that given row.}
	\renewcommand{\tabcolsep}{2.0pt}
    \renewcommand{\arraystretch}{1.05}
    \begin{scriptsize}

\begin{tabular}{ll|c||cc|cc||cc|cc}
  \toprule
High-Level & \multirow{2}{*}{Dim.} & Binary & \multirow{2}{*}{PCA} & PCA- & \multirow{2}{*}{MC} & \multirow{2}{*}{rMC} & \multicolumn{2}{c|}{Transf. (k3)} & \multicolumn{2}{c}{Transf. (k5)} \\
Property & & Relevance &  & Func &  &  & p100 & p500 & p100 & p500 \\ 
  \midrule
\multirow{5}{*}{Multi-} & 2 & \cellcolor{gray!20}0.9973 & 0.9939 & 0.9933 & 0.9738 & 0.9714 & 0.9908 & 0.9971 & 0.9803 & 0.9942 \\ 
  \multirow{5}{*}{modality} & 3 & \cellcolor{gray!20}0.9973 & 0.9603 & 0.9574 & 0.9583 & 0.9540 & 0.9876 & 0.9939 & 0.9709 & 0.9919 \\ 
   & 5 & 0.9989 & 0.8970 & 0.8979 & 0.9602 & 0.9472 & 0.9908 & \cellcolor{gray!20}0.9991 & 0.9883 & 0.9984 \\ 
   & 10 & \cellcolor{gray!20}0.9996 & 0.8386 & 0.8378 & 0.9529 & 0.9522 & 0.9741 & 0.9909 & 0.9694 & 0.9880 \\ 
   \cmidrule{2-11} & all & \cellcolor{gray!20}0.9983 & 0.9211 & 0.9203 & 0.9613 & 0.9562 & 0.9858 & 0.9953 & 0.9771 & 0.9931 \\ 
   \midrule\multirow{5}{*}{Global} & 2 & 0.9969 & 0.9925 & 0.9921 & 0.9579 & 0.9649 & 0.9913 & \cellcolor{gray!20}0.9979 & 0.9811 & 0.9931 \\ 
   \multirow{5}{*}{Structure} & 3 & \cellcolor{gray!20}0.9963 & 0.9515 & 0.9564 & 0.9126 & 0.8985 & 0.9857 & 0.9936 & 0.9660 & 0.9892 \\ 
   & 5 & \cellcolor{gray!20}0.9976 & 0.8067 & 0.8067 & 0.8898 & 0.8586 & 0.9784 & 0.9951 & 0.9736 & 0.9944 \\ 
   & 10 & \cellcolor{gray!20}0.9990 & 0.7737 & 0.7672 & 0.9081 & 0.9112 & 0.9632 & 0.9845 & 0.9534 & 0.9842 \\ 
   \cmidrule{2-11} & all & \cellcolor{gray!20}0.9975 & 0.8713 & 0.8700 & 0.9162 & 0.9067 & 0.9795 & 0.9928 & 0.9684 & 0.9901 \\ 
   \midrule\multirow{5}{*}{Funnel} & 2 & 0.9985 & 0.9991 & 0.9997 & 0.9961 & 0.9955 & 0.9991 & \cellcolor{gray!20}1.0000 & 0.9973 & \cellcolor{gray!20}1.0000 \\ 
   \multirow{5}{*}{Structure} & 3 & 0.9997 & 0.9964 & 0.9961 & 0.9982 & 0.9923 & \cellcolor{gray!20}1.0000 & \cellcolor{gray!20}1.0000 & \cellcolor{gray!20}1.0000 & \cellcolor{gray!20}1.0000 \\ 
   & 5 & \cellcolor{gray!20}1.0000 & 0.9902 & 0.9885 & 0.9967 & 0.9893 & \cellcolor{gray!20}1.0000 & \cellcolor{gray!20}1.0000 & 0.9991 & \cellcolor{gray!20}1.0000 \\ 
   & 10 & \cellcolor{gray!20}1.0000 & 0.9774 & 0.9737 & 0.9955 & 0.9911 & 0.9991 & \cellcolor{gray!20}1.0000 & 0.9994 & \cellcolor{gray!20}1.0000 \\ 
   \cmidrule{2-11} & all & 0.9995 & 0.9907 & 0.9894 & 0.9966 & 0.9921 & 0.9996 & \cellcolor{gray!20}1.0000 & 0.9990 & \cellcolor{gray!20}1.0000 \\ 
   \bottomrule
\end{tabular}
    \end{scriptsize}
    \label{tab:performance_results}
\end{table}

It comes with no surprise to us that the PCA-based approaches perform the worst on high dimensional data. As explained we use LHS for sampling the set of candidate solutions $X$. LHS is optimized to sample a uniformly distributed set of solutions and as the variance in every dimension is similar to one-another, the resulting PCs have also similar variance. Hence, PCA causes in our setting a severe loss of information. Our proposed method \textit{PCA-Func} compensates this effect, slightly. Yet, PCA and PCA-Func both performed better than MC and rMC for $d=[2,3]$.

To our surprise, however, the rMC approach performed rather well for $d=[5,10]$ (contrary to what we expected). Therefore, we conclude that reducing the fitness maps into a single channel by \textit{mean}-aggregation still preserves enough information to identify the considered high-level properties for $d\leq10$. Interestingly, both PCA approaches perform better than the MC and rMC approaches for low dimensions. An explanation may be that the MC approach only receives a single meaningful channel for $d=2$ out of $45$ available channels as the other $44$ channels are empty. This growing sparsity in information may cause the MC approach to underperform for small dimensions. The same reason may apply to the rMC approach. As explained, the various different 2d-projections are aggregated into a single one. Due to this process, the information density within the input image grows for larger dimensionality. 

Contrary, the transformer-based methods perform well over all dimensions. The loss in information for larger dimensions or growing sparsity for smaller ones does not apply here. Instead, they can identify more complex structures (such as global structure) in small and large dimensional spaces, equally well. However, one limitation of the implementation is that the transformer cannot be trained on a dataset with a varying number of sampled points. Yet, the differences in performance between the $np=100$ and $np=500$ models are rather small. So, we do not see any constraint in comparing the transformer model to the other approaches.

Our findings show that our proposed feature-free approaches for high-level property prediction provide competitive performance to the feature-based approaches. 
The main advantages of both, \textit{fitness map} and \textit{fitness cloud}, over the classical approach is (1) their lower computational complexity as no features must be computed, and (2) the absence of domain knowledge about the individual features as well as their selection process. This means, that applying our proposed methods to other optimization tasks is straightforward because the input representation is not tailored to a specific task or problem domain. Next, the \textit{fitness cloud} clearly outperforms the \textit{fitness map} approaches. On the one hand, the number of weights for the transformers' input embedding grows only linear with the number of dimensions in contrast to the MC approach where the number of weights grow with $\binom{d}{2}$. On the other hand, transformers  (and also MC) can only handle data with a limited dimensionality.

Similar, the main advantage of the  classical approach is its independence of (1) the problem-dimensionality, and (2) the used model. In addition, in some cases the interpretation of features is easier than the one of images or especially point clouds. For instance, some of the ELA meta level features, such as the adjusted coefficient of variation are quite easy to understand and interpret -- even for high-dimensional problems.

\section{Conclusion}
\label{sec:conclusion}
In this paper, we proposed several extensions for the \textit{fitness map} approach which was originally proposed by \citet{PragerMP2021TowardsFeatureFree}. With these extensions, the \textit{fitness map} can be applied to high-dimensional data. We could demonstrate that these extensions provide a similar but slightly weaker overall performance than the conventional approach. A major drawback of our proposed image-based techniques is the trade-off between information loss for larger dimensions or growing sparsity for smaller ones. Therefore, we proposed another, 3d-point cloud-based approach. This approach can be considered as highly competitive to the conventional approach.

In future work, we want to use these proposed techniques to extend and improve on recent work on feature-free AAS, e.g.,~\cite{alissa2019algorithm,seiler2020deep, PragerMP2021TowardsFeatureFree}. 
As the discussed high-level properties are well-known to determine problem hardness, and AAS studies revealed that different types of algorithms are suitable candidates for respective problem classes, sequential algorithm selection models could also be promising alternatives. By this means, the candidate solver portfolio could be flexibly restricted to a promising subset, which is well-suited for the type of the underlying problems, based on the high-level properties as determined by the ELA features or feature-free alternatives presented here. 
Next, we want to explore the potential of \textit{dynamic} AAS in the context of 
continuous optimization problems and test our proposed approach on real-world problems. Further, we want to explore the potential of a learn-able version of our rMC approach. Instead of simple mean aggregation, we plan to use a $1\times1$ conv-layer to project the $n$ feature maps into a single one.




\bibliographystyle{unsrt}  
\bibliography{references}  
\end{document}